# SalNet360: Saliency Maps for omni-directional images with CNN

Rafael Monroy*, Sebastian Lutz*, Tejo Chalasani*, Aljosa Smolic

{*monroyrr,lutzs,chalasat,smolica*}*@scss.tcd.ie*

*Trinity College Dublin, Ireland*

**Abstract**

The prediction of Visual Attention data from any kind of media is of valuable use to content creators and used to efficiently drive encoding algorithms. With the current trend in the Virtual Reality (VR) field, adapting known techniques to this new kind of media is starting to gain momentum. In this paper, we present an architectural extension to any Convolutional Neural Network (CNN) to fine-tune traditional 2D saliency prediction to Omnidirectional Images (ODIs) in an end-to-end manner. We show that each step in the proposed pipeline works towards making the generated saliency map more accurate with respect to ground truth data.

## 1. Introduction

The field of Virtual Reality (VR) has experienced a resurgence in the last couple of years. Major corporations are now investing considerable efforts and resources to deliver new Head-Mounted Displays (HMDs) and content in a field that is starting to become mainstream.

Displaying Omni-Directional Images(ODIs) is an application for VR headsets. These images portray an entire scene as seen from a static point of view, and when viewed through a VR headset, allow for an immersive user experience. The most common method for storing ODIs is by applying equirectangular, cylindrical or cubic projections and saving them as standard two-dimensional images [1].

---

*These authors contributed equally to this work.

One of the many directions of research in the VR field is Visual Attention, the main goal of which is to predict the most probable areas in a picture the average person will look at, by analysing an image. As shown by Rai el at. [2], visual attention is the result of two key factors: bottom-up saliency and top-down perceived. In order to collect ground truth data, experiments are performed in which subjects look at pictures while an eye-tracker, together with the Inertial Measurement Unit of the headset in use, records the location in the image the user is looking at [3]. By collecting this data from several subjects, it is possible to create a saliency map that highlights the regions where most people looked at. Knowing which portions in an image are the most observed can be used, for example, to drive compression and segmentation algorithms [4][5].

Earlier works on image saliency made use of manually-designed feature maps that relate to salient regions and when combined in some form produce a final saliency map [6][7][8]. With the advent of deep neural networks, GPU-optimised code and availability of vast amounts of annotated data, research efforts in computer vision tasks such as recognition, segmentation and detection have been focused on creating deep learning network architectures to learn and combine features maps automatically driven by data [9][10][11]. As evident from the MIT Saliency Benchmarks [12], variants of Convolutional Neural Networks (CNNs) are the top-performing algorithms for generating saliency maps in traditional 2D images. Good techniques for transfer learning[13] made sure that CNNs can be used for predicting user gaze even with a relatively small amount of data [14][15][16][17].

Saliency prediction techniques originally designed for traditional 2D images cannot be directly used on ODIs due to the heavy distortions present in most projections and a different nature in observed biases. De Abreu et al. [18] demonstrated that removing the centre bias present on most techniques for traditional 2D images significantly improves the performance when estimating saliency maps for ODIs.

A recent effort to attract attention to the problem of creating saliency maps for ODIs was presented in the Salient360! Grand Challenge at the ICME 2017 Conference[19]. Participants were provided with a series of ODIs (40 in total) for which head- and eye-tracking ground truth data was given in the form of saliency maps. The details on how the dataset was built are described by Rai et al. [3]. Twenty-five paired images and ground truth were later provided to properly evaluate the submissions to the challenge. In the presented paper we follow the experimental conditions of the challenge, i.e. we used the initial 40 images to train the proposed CNN and the results we present were calculated using the 25 test images.

The approach we present is similar to that of De Abreu et al. [18] in that the core CNN to create the saliency maps can be switched once better networks become available. We start with the premise that the heavy distortions near the poles in an equirectangular ODI negatively affect the final saliency map and the nature of the biases near the equator and poles differs from those in a traditional 2D image. To address these issues, we make the following contributions:

- Subdividing the ODI into undistorted patches.

- Providing the CNN with the spherical coordinates for each pixel in the patches.

This paper is organised as follows. Section 2 enlists work related to saliency maps for traditional 2D images and ODIs. Then, Section 3 describes the pipeline and the required pre- and post-processing steps. The end-to-end trainable CNN architecture we use here is then described in Section 4. Results are presented in Section 5. Finally, in Section 6 ideas for future research opportunities and conclusions are provided.

## 2. Previous work

Until the advent of CNNs, most saliency models relied on extracting feature maps, which depict the result of decomposing the image, highlighting a particular characteristic. These feature maps are then linearly combined by computing a series of weights to be applied to each map. Features like those presented by Itti and Koch [6] were initially based on knowledge of the low-level human vision, e.g. color, orientation and frequency responses.

The catalogue of features used to calculate saliency maps has increased over the years, including for example, global features [7], face detection features [20], location bias features [21] and others. Judd et al.[8] created a framework and model that combined many of these features to compute saliency maps.

The MIT Saliency Benchmark was later developed by Bylinskii et al., allowing researchers to submit and compare their saliency computation algorithms [12].

With the increasing availability of VR headsets, saliency computation methods specifically tailored for ODIs have started to surface. Due to the fact that ODIs describe, in actuality, a sphere, saliency models designed for traditional 2D images cannot be directly used. Bogdanova et al. analyse the sphere depicted in an ODI, creating low-level vision feature maps based on intensity, colour and orientation features [22].

In recent years, the interest in CNNs has grown and most of the top-performing submissions in the MIT Saliency Benchmark system correspond to CNN-based approaches. One of the first methods to use CNNs for saliency was presented by Vig et. al. [23], which used feature maps from convolution layers as the final features of a linear classifier. Kümmerer et. al. [13] published an extension of this approach to use CNNs as feature extractors. They extended their approach in [24], where they use the VGG [25] features to predict saliency with no fine-tuning. Instead they only train a few readout layers on top of the VGG layers. The method developed by Liu et. al. [14] used a Multi-Resolution CNN, where the network was trained with image regions centred on fixation and non-fixation eye locations at several scales. Another approach using CNNs to extract features was shown by Li et. al. [26], where they obtained features from convolutional layers at several scales and combined them with handcrafted features which were trained on a Random Forest. One of the first end-to-end trained CNNs for saliency prediction was introduced by Pan et. al. [17]. Similar work from Kruthiventi et al. [16] added a novel Location Biased Convolutional Layer, which allowed them to model location dependent patterns.

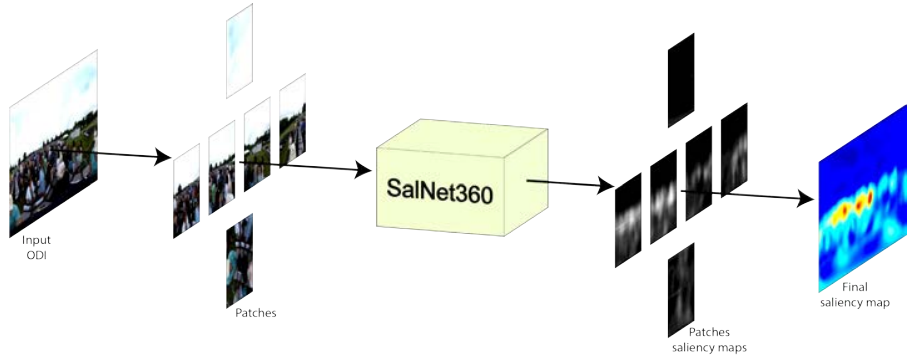

Figure 1: ODI Saliency Detection Pipeline.

As CNNs are becoming more complex and sophisticated in other Computer Vision domains, these advances are being transferred to saliency prediction applications, for example, the method proposed by Pan et. al. [27], introducing adversarial examples to train their network. The work of Liu et. al. [28] introduces a deep spatial contextual LSTM to incorporate global features in their saliency prediction.

The amount of published research focusing on CNNs applied to ODIs is at the moment very small. De Abreu et. al. [18] uses translated versions of the ODI as input to a CNN trained for traditional 2D images to generate an omnidirectional saliency map.

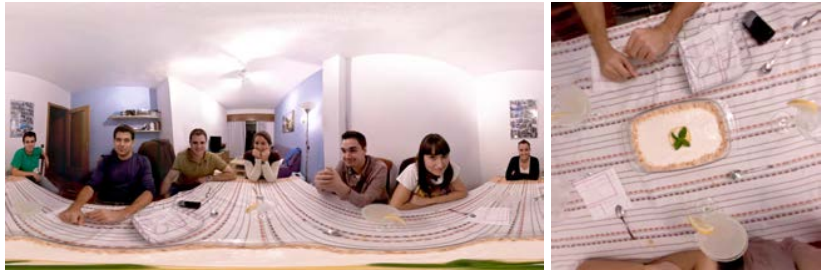

(a) Full ODI      (b) Nadir view

Figure 2: Example of distortions present in equirectangular ODIs.

## 3. Method

In this section we detail the pipeline used to obtain the saliency map for a given ODI. Fig. 1 shows a diagram with the complete pipeline. Our method takes an ODI as input and splits it into six patches using the pre-processing steps described in 3.1. Each of these six patches is sent through the CNN, the details of which we delve into in Section 4. The output of the CNN for all the patches are then combined using the post-processing technique mentioned in Section 3.2

### 3.1. Pre-processing

Mapping a sphere onto a plane requires introducing heavy distortions to the image, which are most evident at the poles. When looking at those distorted areas in the ODI, it becomes very hard to identify what they are supposed to be depicting. Fig. 2 gives an example of the distortions observed on an equirectangular ODI. From Fig. (2a) alone, it is not possible to recognise the object on the table. Once the nadir view is undistorted as seen in Fig. (2b), it is clear that the object in question is a cake.

In order to reduce the effect of these distortions on saliency estimation, we divide the ODI into equally-sized patches by rendering viewing frustums with a field of view of (FOV) approximately 90 degrees each. This field of view was selected to keep distortions low, cover the entire sphere with six patches and have a similar FOV to that of the Oculus Rift, which was used to create the dataset (approximately 100 degrees).

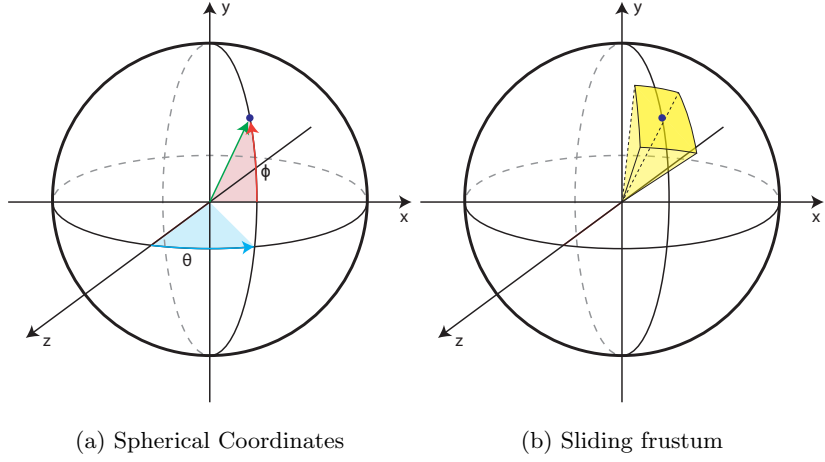

(a) Spherical Coordinates  (b) Sliding frustum

Figure 3: Spherical coordinates definition and sliding frustum used to create the patches.

By specifying the field of view per patch and its resolution, it is possible to calculate the spherical coordinates of each pixel in the patch. These are then used to find the corresponding pixels in the ODI by applying the following equations:

$$x = s_w \left( \frac{\theta + \frac{\pi}{2}}{2\pi} \right) \qquad (1)$$

$$y = s_h \left( 1 - \frac{\phi + \frac{\pi}{2}}{\pi} \right) \qquad (2)$$

Where $\theta$ and $\phi$ are the spherical coordinates of each pixel, see Fig. 3a. The variables $s_w$ and $s_h$ correspond to the ODI's width and height respectively. A graphical representation of how patches look like when sampling the sphere can be seen in Fig. 3b.

The process of generating patches is also applied during training, which will be discussed in Section 4.2. During the saliency map computation six patches with fixed views are generated. Two of these views are oriented towards the nadir and zenith, the other four are pointed towards the horizon but rotated horizontally to cover the entire band at the sphere's equator. Fig. 4 illustrates such partitions when applied to an equirectangular ODI. Though one could think of these views as a cube map, our definition is more generic in the sense that the FOV can be adjusted to be closer to that of the device used to visualise the ODI, or increase the number of patches to further decrease the distortions.

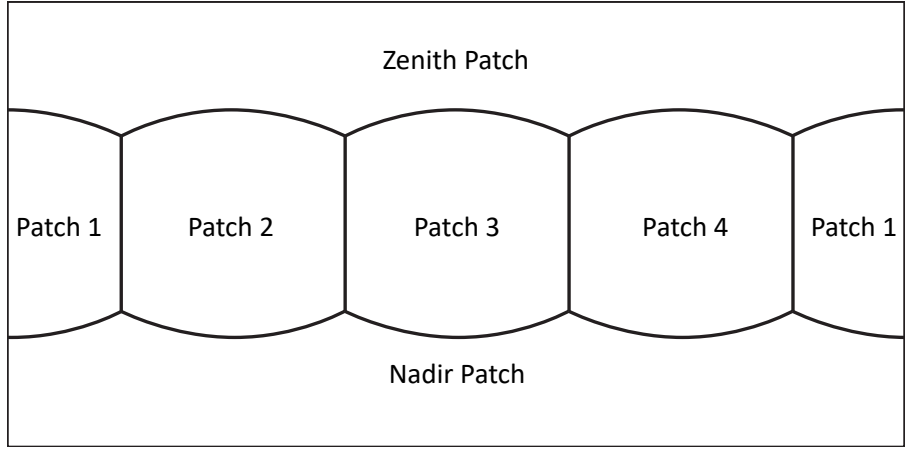

Figure 4: Patches extracted from the ODI.

3.2. Post-processing

As previously mentioned, the CNN takes the six patches and their spherical coordinates as inputs. As result, the CNN generates a saliency map for each of these patches. Consequently, they have to be combined to a single saliency map as output. For that, we project each pixel of each patch to the equirectangular ODI, using their per-pixel spherical coordinates. We apply forward-projection with nearest-neighbour interpolation for simplicity. In order to fill holes and smooth the result, we apply a Gaussian filter with a kernel size of 64 pixels. This kernel size was selected because it was found to give the best results in terms of correlation with the ground truth. Such processing is efficient and sufficient, as we do not compute output images for viewing, but estimate saliency maps.

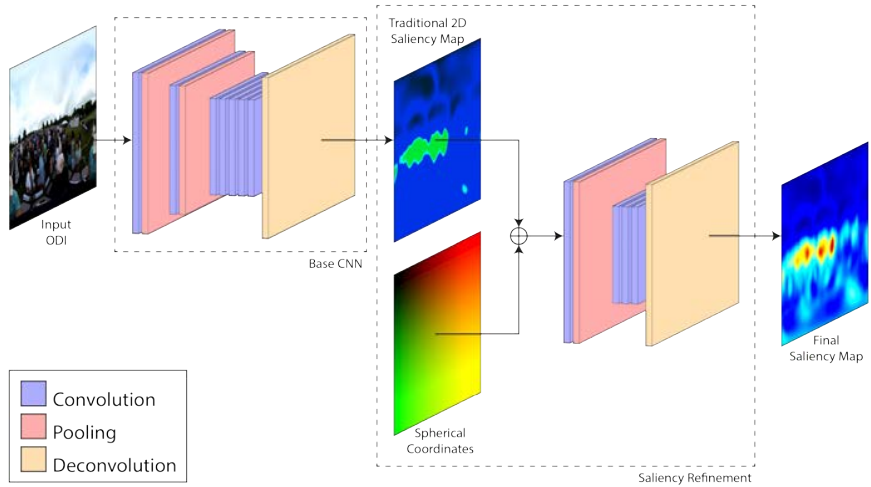

Figure 5: Network Architecture.

## 4. SalNet360

Since ODIs depict an entire 360-degree view, their size tend to be considerably large. This factor, together with the current hardware limitations, prohibits using them directly as inputs in a CNN without heavily down-scaling the ODI. Deep CNNs need large amounts of data to avoid over-fitting. Since training data consisted of only 40 images, the resulting amount of data would not be enough for properly training a CNN.

To address this problem we generated one hundred patches (paired colour image and ground truth saliency map) per ODI by placing the viewing frustum to random locations. The saliency maps per patch are used as labels in the CNN. Fig. 5 illustrates the architecture of the proposed network.

Our network consists of two parts, the Base CNN and the refinement architecture. The Base CNN is trained to detect saliency maps for traditional 2D images. It has been pre-trained using the SALICON dataset [29]. The second part is a refinement architecture that is added after the Base CNN. It takes a 3-channel feature map as input: the output saliency map of the Base CNN and the spherical coordinates per pixel as two channels. This combination of the Base CNN and the Saliency Refinement has been trained as will be discussed in Section 4.2. Omnidirectional saliency maps differ from traditional 2D saliency maps in that they are affected by the view or position of the users head. Combining the spherical coordinates of each pixel as extra input provides the network in the second stage the information to highlight or lower the already computed salient regions depending on their placement in the ODI. We delve into the details of network architecture and training in the next subsections.

Table 1: Network parameters.

| Layer | Input depth | Output depth | Kernel size | Stride | Padding | Activation |
|---|---|---|---|---|---|---|
| conv1 | 3 | 96 | 7×7 | 1 | 3 | ReLU |
| pool1 | 3 | 3 | 3×3 | 2 | 0 | - |
| conv2 | 96 | 256 | 5×5 | 1 | 2 | ReLU |
| pool2 | 256 | 256 | 3×3 | 2 | 0 | - |
| conv3 | 256 | 512 | 3×3 | 1 | 1 | ReLU |
| conv4 | 512 | 256 | 5×5 | 1 | 2 | ReLU |
| conv5 | 256 | 128 | 7×7 | 1 | 3 | ReLU |
| conv6 | 128 | 32 | 11×11 | 1 | 5 | ReLU |
| conv7 | 32 | 1 | 13×13 | 1 | 6 | ReLU |
| deconv1 | 1 | 1 | 8×8 | 4 | 2 | - |
| merge | - | - | - | - | - | - |
| conv8 | 3 | 32 | 5×5 | 1 | 2 | ReLU |
| pool3 | 32 | 32 | 3×3 | 2 | 0 | - |
| conv9 | 32 | 64 | 3×3 | 1 | 2 | ReLU |
| conv10 | 64 | 32 | 5×5 | 1 | 2 | ReLU |
| conv11 | 32 | 1 | 7×7 | 1 | 3 | ReLU |
| deconv2 | 1 | 1 | 4×4 | 2 | 1 | - |

4.1. Network Architecture

The architecture of our Base CNN was inspired by the deep network introduced by Pan et. al. [17] and shares the same layout as the VGG_CNN_M architecture from [30] in the first three layers. This allowed us to initialise the weights of these layers with the weights that have been learned on the ImageNet classification task. After the first three layers, four more convolution layers follow, each of them followed by a Rectified Linear Unit (ReLU) activation function. Two max-pooling layers after the first and second convolution layers reduce the size of the feature maps for subsequent convolutions and add some translation invariance. To upscale the final feature map to a saliency map with the same dimensions as the input image, a deconvolution layer is added at the end. As will be discussed in Section 4.2, we train our whole network in two stages. In the first stage only this Base CNN is trained on traditional 2D images with the help of an Euclidean Loss function directly after the deconvolution layer. In our final network, however, this loss function is moved to the end of our refinement architecture and the output of the deconvolution layer is merged with the 2-channel per-pixel spherical coordinates as input to the second part of our architecture, the Saliency Refinement.

As has been stated above, our architecture does not use the entire ODI at the same time. Instead, the ODI is split into patches and for each of these patches the saliency map is calculated individually. The idea of the second part of our architecture, the Saliency Refinement, is to take the saliency map generated from the Base CNN and refine it with the information on where in the ODI this patch is located. For example, this allows the model to infer the strong bias towards the horizon that the ground truth saliency maps for ODIs show. The whole Saliency Refinement stage consists of four convolution layers, one max-pooling layer after the first convolution and one deconvolution layer at the end. All activation functions are ReLUs and the loss function at the end is Euclidean Loss. Our full architecture is visualised in Fig. 5 and the hyperparameters for all layers are presented in Table 1.

We experimented with different activation layers and different loss functions in the network, however, we found that using ReLUs and Euclidean Loss yielded the best metrics on the dataset.

*4.2. Training*

We performed training in two stages. In the first stage we only train the first part of the network, the Base CNN. The first three layers are initialised using pre-trained weights from VGG_CNN_M network from [30]. We then use SALICON data [29] to train the first part of the network. The data (both images and their saliencies) is normalised by removing the mean of pixel intensity values and re-scaling the data to a [-1,1] interval. We also scale all the images and saliency maps to 360x240 resolution and split the dataset into two sets of 8000 images for training and 2000 images for testing. The network is then trained for 20,000 iterations using the stochastic gradient descent method and with batch size of four.

For the second stage of the training we add the Saliency Refinement part to the network and use the dataset we created with ODIs from [3]. As described in Section 3.1, 100 patches are randomly sampled per ODI to generate a dataset of 4000 images, their ground truth saliency and spherical coordinates. This data augmentation strategy allows us to train the network even though we only had a few ODIs available. We pre-process all the images using the same techniques we used for the first stage of training. The weights for the Base CNN are initialised using the weights we obtained from the first stage training. We started with a base learning rate of 1.3e-7 and reduced it by 0.7 after every 500 iterations. The network is trained for 22,000 iterations with a 10% split for test data and the rest of the images are used for training. The batch size is set to be five while the test error is monitored every 100 iterations to look for divergence. To avoid overfitting we additionally use standard weight decay as regularization. We present our experiments and results in the next section.

## 5. Results

In this section we describe the experiments and results that indicate that our method enhances a CNN trained with traditional 2D images and allows it to be applied to ODIs. Before we discuss the actual numbers, we provide a short introduction to each of the metrics used to evaluate the generated saliency maps.

### 5.1. Performance measures

Bylinskii et al. [31] differentiate between two types of performance metrics based on their required ground truth format. Location-based metrics need a ground truth with values at discrete fixation locations, while distribution-based metrics consider the ground truth and saliency maps as continuous distributions.

Four metrics were used to evaluate the results of our system: the Kullback-Leibler divergence (KL), the Pearson's Correlation Coefficient (CC), the Normalized Scanpath Saliency (NSS) and the Area under ROC curve (AUC). Both the KL and the CC are distribution-based metrics, whereas the NSS and AUC are location-based. In the case of the former two metrics, the predicted saliency map of our approach is normalised to generate a valid probability distribution. All these metrics were designed with traditional 2D images in mind. Gutiérrez et al. [32] developed a toolbox that is specifically tailored to analyse ODIs. The results we present were obtained using these new tools. We refer to Gutiérrez et al. work for further details.

5.2. Experiments

In order to test the improvements obtained from our proposed method, we defined three scenarios to be compared. The first one consists of applying our Base CNN, see Fig. 5 while taking the entire ODI as input by downscaling it to a resolution of 800×400 pixels. The predicted saliency map is then upscaled to the resolution of the original image. In our second scenario, we divide the ODI in six undistorted patches as described in Section 3.1 and run these patches separately through the Base CNN. Afterwards we recombine the predicted saliencies of these patches to form the final saliency map. Finally, the third scenario consists of the whole pipeline as described in Section 3, where the spherical coordinates of the patches are also considered by the CNN during inference.

The results of our experiments can be seen in Table 2, where we show the average KL, CC, NSS and AUC for the 25 test images. As can be seen from the table, only running the entire ODI through the Base CNN gives relatively poor results compared to the later scenarios. By dividing the ODI into patches and using them individually before being recombined, our results are improved in most of the metrics. After adding in the spherical coordinates to the inference of each patch, the results clearly improve considerably in all the metrics.

A visual example of the results of the three scenarios can be seen in figure 6. In the top row from left to right, the input ODI and a blended version of the ODI and ground truth are shown. In the bottom row again from left to right the predicted saliency maps blended into the input ODI for each scenario are presented. The first scenario on the far left (Base CNN only) predicted two big salient centres, which are not found in the ground truth map. The second scenario improved on these predictions by finding salient areas that also cover salient areas in the ground truth. However, it still predicts two highly salient areas in wrong locations and introduced some salient areas of low impact at the bottom part of the image that are incorrectly labelled as salient. Finally, the third scenario clearly covers all the salient areas in the ground truth and removes the incorrectly labelled salient regions at the bottom of the image that the second scenario introduced. It is, however, too generous in the amount of area that the salient regions cover, which seems to be the main issue in this scenario.

Table 2: Comparison of the three experimental scenarios.

∗ indicates a significant improvement in performance compared to the Base CNN (t-test, $p < 0.01$).

|                          | KL     | CC     | NSS    | AUC    |
|--------------------------|--------|--------|--------|--------|
| Base CNN                 | 1.597  | 0.416  | 0.630  | 0.648  |
| Above + Patches          | 0.625  | 0.474  | 0.566  | 0.659  |
| Above + Spherical Coords.| 0.487* | 0.536* | 0.757* | 0.702* |

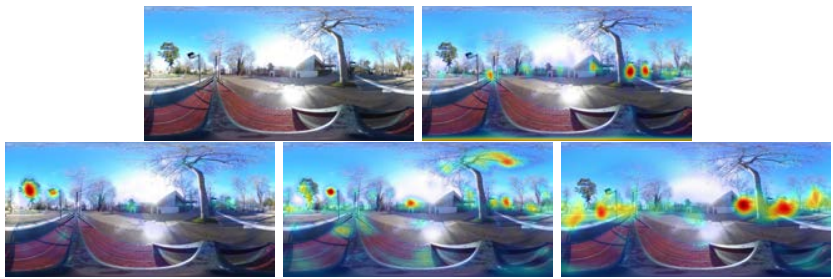

Figure 6: Comparison of the three experimental scenarios. *Top row:* On the left the input ODI, on the right the ground truth saliency map blended with the image.
*Bottom row:* From left to right, the result of the three experimental scenarios: Base CNN, Base CNN + Patches, Base CNN + Patches + Spherical Coords.

In Fig. 8, some of the best-performing results are shown. As a compari-son, two of the lowest-performing results are shown in Fig. 9. A summary of the results of all images can be seen in Appendix A. Fig. 7 provides a visual representation of the values obtained for each metric.

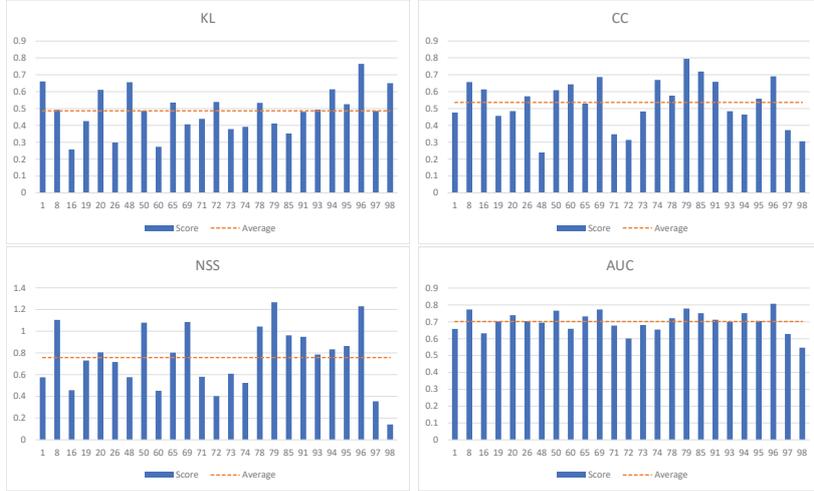

Figure 7: Plots for each of the metrics applied to the test ODIs.

5.3. Salient360! Grand Challenge

As mentioned in Section 1, a challenge was organised during the ICME 2017 Conference, in which participants were provided with training data consisting of 40 ODIs paired with head- and eye-tracking ground truth data. Participants were allowed to submit their work on three different categories: *head*, *head+eye* and *scanpath*. The goal of the first category being the estimation of a saliency map considering only the orientation of the head; the objective of the second category corresponded to the estimation of a similar saliency map but adding eye-tracking information, leading to more localised predictions; finally the third one had the goal of estimating a collection of scanning paths that would be compared to the scanning paths collected by the organisers. The work here presented was submitted and participated in the second category: *head+eye*.

A total of 16 individual submissions were evaluated in the *head+eye* category. Table 3 shows the results for the top-5 performers in the challenge from the same

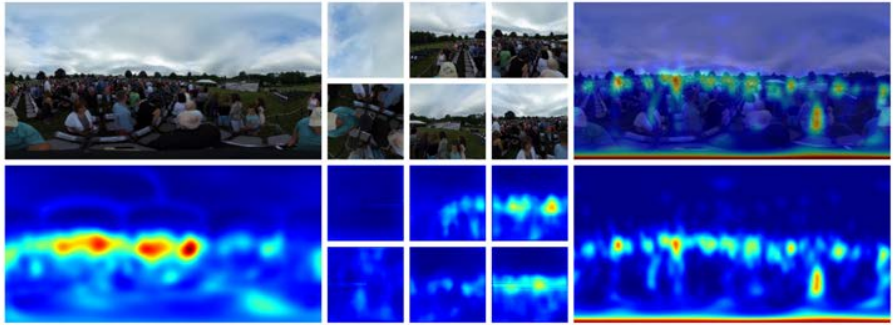

(a) Image index: 69, KL: 0.433, CC: 0.686, NSS: 1.064, AUC: 0.769

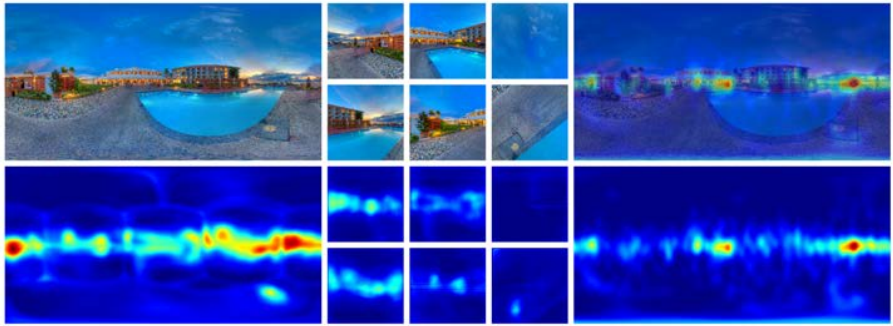

(b) Image index: 79, KL: 0.360, CC: 0.818, NSS: 1.303, AUC: 0.779

Figure 8: Two of the best-performing examples. *Top row:* From left to right: Input image, six extracted patches, ground truth blended with input image. *Bottom row:* From left to right: Predicted saliency map, predicted saliency maps from each patch, ground truth saliency map.

institution.

We demonstrated that augmenting the CNN's input with the corresponding spherical coordinates and using undistorted patches of the ODI leads to better saliency predictions. This approach could easily be applied to more accurate predictors trained for traditional 2D images.

*5.4. Challenges*

As can be seen on the predicted saliency maps in Figs. 8 and 9, the merging of the predicted patches leaves a distinct pattern in the final saliency map, leaving in some cases a lattice-like pattern. We tried to mitigate this issue by applying a Gaussian Blur, but it is still noticeable in some of the results and consequently has a negative effect on the *KL* and *CC* scores. In the saliency prediction of the individual patches in both figures, another set of artefacts can also be observed. We speculate that these line patterns stem from the addition of the spherical coordinates, because these predictions correspond to the patches that show the poles. However, when recombining the patches and applying the post-processing steps discussed above, these artefacts are no longer noticeable. We expect that increasing the amount of training data could help alleviate some of the issues.

Table 3: Top-5 performers in the challenge

| Model name | KL | CC | NSS | AUC |
|---|---|---|---|---|
| TU Munich [33] | 0.449 | 0.579 | 0.805 | 0.726 |
| SJTU [34] | 0.481 | 0.532 | 0.918 | 0.735 |
| Wuhan [35] | 0.508 | 0.538 | 0.936 | 0.736 |
| ProSal / Zhejihang [36] | 0.698 | 0.527 | 0.851 | 0.714 |
| **SalNet360 (ours)** | **0.487** | **0.536** | **0.757** | **0.702** |

## 6. Conclusions

We showed in Section 5, that dividing an omnidirectional image into patches and adding a Saliency Refinement architecture that takes into consideration spherical coordinates to an existing Base CNN can considerably improve the results in omnidirectional saliency prediction. We envision several potential improvements that could be made to increase performance. Our network was trained using the Euclidean Loss function, which is a relatively simple loss function that is applicable to a wide variety of cases, such as saliency prediction. It tries to minimise the pixel-wise difference between the calculated saliency map and the ground truth. However, to optimise based on any of the performance metrics mentioned in 5.1, custom loss functions could be used. In this way, the network would be trained to specifically minimise in regards to the *KL*, *CC* or *NSS*.

As mentioned in Section 5.4, one of the bigger issues that affect our results are the artefacts that are created when recombining the patches to create the

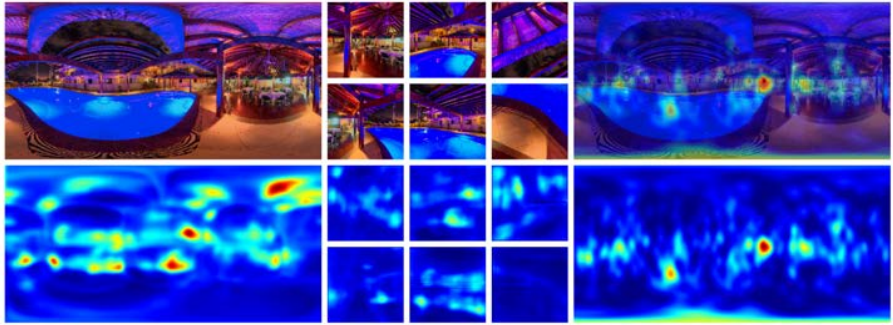

(a) Image index: 72, KL: 0.459, CC: 0.271, NSS: 0.353, AUC: 0.609

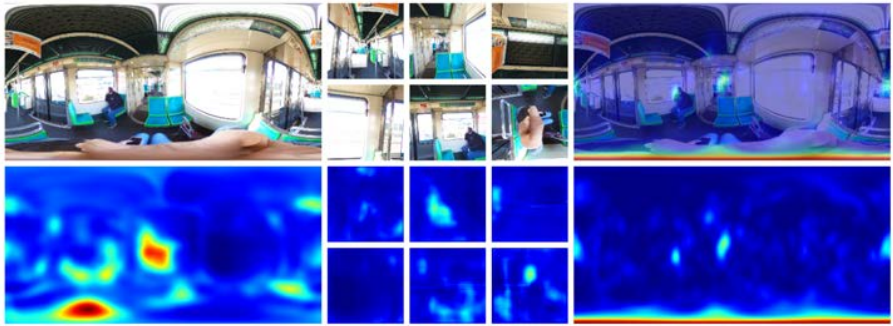

(b) Image index: 98, KL: 0.652, CC: 0.309, NSS: 0.126, AUC: 0.529

Figure 9: Two of the lowest-performing examples. *Top row:* From left to right: Input image, six extracted patches, ground truth blended with input image. *Bottom row:* From left to right: Predicted saliency map, predicted saliency maps from each patch, ground truth saliency map.

final saliency map. Instead of just using Gaussian Blur to remove these artefacts, a more sophisticated process could be implemented since these artefacts always appear in the same way and location.

In most Deep Learning applications, improvements are often made by making the networks deeper. Our network is based on the Deep Convolutional Network by Pan et. al. [17], but compared to the current state-of-the-art in other Computer Vision tasks e.g. classification and segmentation, this network is still relatively simple. We are confident that updating the Base CNN to recent advances will improve the results in omnidirectional saliency.

Finally, as with all Deep Learning tasks, a large amount of data is needed to achieve good results. Unfortunately there is currently only a relatively small amount of ground truth saliency maps available, and even less data for omnidirectional saliency. It is our hope that more data will become available in the future which can be used to get better results.

In this work, we present an end-to-end CNN that is specifically tailored for estimating saliency maps for ODIs. We provide evidence which indicates that part of the discrepancies found when using CNNs trained on traditional 2D images is due to the heavy distortions found on the projected ODIs. These issues can be addressed by dividing the ODI in undistorted patches before calculating the saliency map. Furthermore, in addition to the patches, biases due to the location of objects on the sphere can be considered by using the spherical coordinates of the pixels in the patches before computing the final saliency map.

**Acknowledgement**

The present work was supported by the Science Foundation Ireland under the Project ID: 15/RP/2776 and with the title V-SENSE: Extending Visual Sensation through Image-Based Visual Computing.

## Appendix A  Summary of results per image

Individual results for each ODI. The two best and worst results have been highlighted in green and red respectively.

| Image index | KL | CC | NSS | AUC |
|---|---|---|---|---|
| 1 | 0.661 | 0.476 | 0.576 | 0.658 |
| 8 | 0.492 | 0.657 | 1.105 | 0.773 |
| 16 | 0.257 | 0.613 | 0.457 | 0.632 |
| 19 | 0.425 | 0.456 | 0.731 | 0.702 |
| 20 | 0.611 | 0.485 | 0.806 | 0.739 |
| 26 | 0.299 | 0.573 | 0.717 | 0.705 |
| 48 | 0.656 | 0.239 | 0.578 | 0.695 |
| 50 | 0.486 | 0.609 | 1.078 | 0.766 |
| 60 | 0.273 | 0.643 | 0.452 | 0.659 |
| 65 | 0.535 | 0.529 | 0.803 | 0.732 |
| 69 | 0.407 | 0.687 | 1.085 | 0.773 |
| 71 | 0.439 | 0.347 | 0.581 | 0.678 |
| 72 | 0.539 | 0.314 | 0.403 | 0.601 |
| 73 | 0.379 | 0.482 | 0.609 | 0.681 |
| 74 | 0.391 | 0.670 | 0.524 | 0.654 |
| 78 | 0.533 | 0.576 | 1.043 | 0.721 |
| 79 | 0.411 | 0.795 | 1.266 | 0.778 |
| 85 | 0.352 | 0.719 | 0.962 | 0.751 |
| 91 | 0.481 | 0.658 | 0.949 | 0.713 |
| 93 | 0.493 | 0.484 | 0.786 | 0.700 |
| 94 | 0.614 | 0.464 | 0.834 | 0.751 |
| 95 | 0.526 | 0.558 | 0.864 | 0.706 |
| 96 | 0.765 | 0.690 | 1.230 | 0.807 |
| 97 | 0.486 | 0.372 | 0.356 | 0.628 |
| 98 | 0.650 | 0.306 | 0.142 | 0.547 |
| **Mean** | **0.487** | **0.536** | **0.757** | **0.702** |
| **Std. Dev.** | **0.128** | **0.144** | **0.290** | **0.061** |